\documentclass[letterpaper, 10 pt, conference]{ieeeconf}  % Comment this line out if you need a4paper

\IEEEoverridecommandlockouts                              % This command is only needed if 
\usepackage[noadjust]{cite}
\usepackage{graphics} % for pdf, bitmapped graphics files
\usepackage{graphicx}
\usepackage[tight,footnotesize]{subfigure}
\usepackage{textcomp}%, libertine} 

\usepackage{mathtools, hyperref}
\usepackage{hyphenat}
\usepackage{xcolor}%tmp
\long\def\invis#1{}

\def\floatsep{4.0pt}
\def\textfloatsep{0.0pt}
\def\dblfloatsep{4.0pt}
\def\dbltextfloatsep{4.0pt}

\title{\LARGE \bf
Efficient Multi-Robot Coverage of a Known Environment
}

\author{
\authorblockN{Nare Karapetyan$^{1,2}$, Kelly Benson$^{1}$, Chris McKinney$^{1}$, Perouz Taslakian$^{2,3}$ and Ioannis Rekleitis$^{1}$\\\medskip
\authorblockA{$^{1}$University of South Carolina, Columbia SC, USA, \\\medskip
{\tt\small nare@email.sc.edu, yiannisr@cse.sc.edu}}
\authorblockA{$^{2}$American University of Armenia,
Yerevan, Armenia,}
\authorblockA{$^{3}$Element AI, Montreal, Canada, \\\medskip
{\tt\small perouz@elementai.com}}
}}

\begin{document}

\maketitle
\thispagestyle{empty}
\pagestyle{empty}

%%%%%%%%%%%%%%%%%%%%%%%%%%%%%%%%%%%%%%%%%%%%%%%%%%%%%%%%%%%%%%%%%%%%%%%%%%%%%%%%
\begin{abstract}
%%%%%%%%%%%%%%%%%%%%%%%%%%%%%%%%%%%%%%%%%%%%%%%%%%%%%
%%                 Abstract                        %%
%%%%%%%%%%%%%%%%%%%%%%%%%%%%%%%%%%%%%%%%%%%%%%%%%%%%%
This paper addresses the complete area coverage problem of a known environment by multiple-robots. Complete area coverage is the problem of moving an end\hyp effector over all available space while avoiding existing obstacles. In such tasks, using multiple robots can increase the efficiency of the area coverage in terms of minimizing the operational time and increase the robustness in the face of robot attrition. Unfortunately, the problem of finding an optimal solution for such an area coverage problem with multiple robots is known to be NP-complete. 
In this paper we present two approximation heuristics for solving the multi\hyp robot coverage problem. The first solution presented is a direct extension of an efficient single robot area coverage algorithm, based on an exact cellular decomposition. The second algorithm is a greedy approach that divides the area into equal regions and applies an efficient single-robot coverage algorithm to each region. 
We present experimental results for two algorithms. Results indicate that our approaches provide good coverage distribution between robots and minimize the workload per robot, meanwhile ensuring complete coverage of the area.

\end{abstract}

%%%%%%%%%%%%%%%%%%%%%%%%%%%%%%%%%%%%%%%%%%%%%%%%%%%%%%%%%%%%%%%%%%%%%%%%%%%%%%%%
\section{INTRODUCTION}
%\section{Introduction}

The problem of coverage is common to many application domains. From household robotics, to automation in agriculture, search and rescue, environmental monitoring, and humanitarian de\hyp mining \cite{Choset-2001, Galceran-2013}. Formally, the \emph{coverage problem} is defined as the planning of a trajectory for a mobile robot such that an end\hyp effector, often the body of the robot, visits every point in the specified available space while avoiding existing obstacles. Utilizing multiple robots can potentially reduce the coverage time, and, in hazardous situations, such as de\hyp mining, increase robustness by completing the mission even in the event of accidental ``robot deaths''. Meanwhile, the utilization of multi-robot teams increases the algorithmic complexity and logistic management. 

Rekleitis et al.~\cite{Rekleitis-2008} showed that finding a minimum-cost traveling path for multi-robot systems is NP-hard,  while for single robot coverage there exists an efficient polynomial-time algorithm~\cite{Xu_et_al.-2014} based on the well-studied Chinese Postman Problem (CPP)~\cite{cpp}.  The current paper extends the work of Xu et al.~\cite{Xu_et_al.-2014} on efficient coverage of a known arbitrary environment by a single robot to coverage by multiple robots. 

We define optimal coverage as a \emph{MinMax} problem: minimize the length of the longest coverage path over a group of robots.  Such a formulation, as opposed to minimizing the overall length of paths, helps us reduce the probability of situations when one or more robots remain idle while the rest do all the work.  It is worth noting that the Chinese Postman Problem, utilized by Xu et al.~\cite{Xu_et_al.-2014}, becomes NP-hard for the case of multiple postmen~\cite{Frederickson:1976}  when the objective is to minimize the maximum path length.

\begin{figure}[t]
  \centering
  \includegraphics[width=0.4\textwidth, clip=true, trim=0.0in 0.0in 0.0in 0.0in]
  {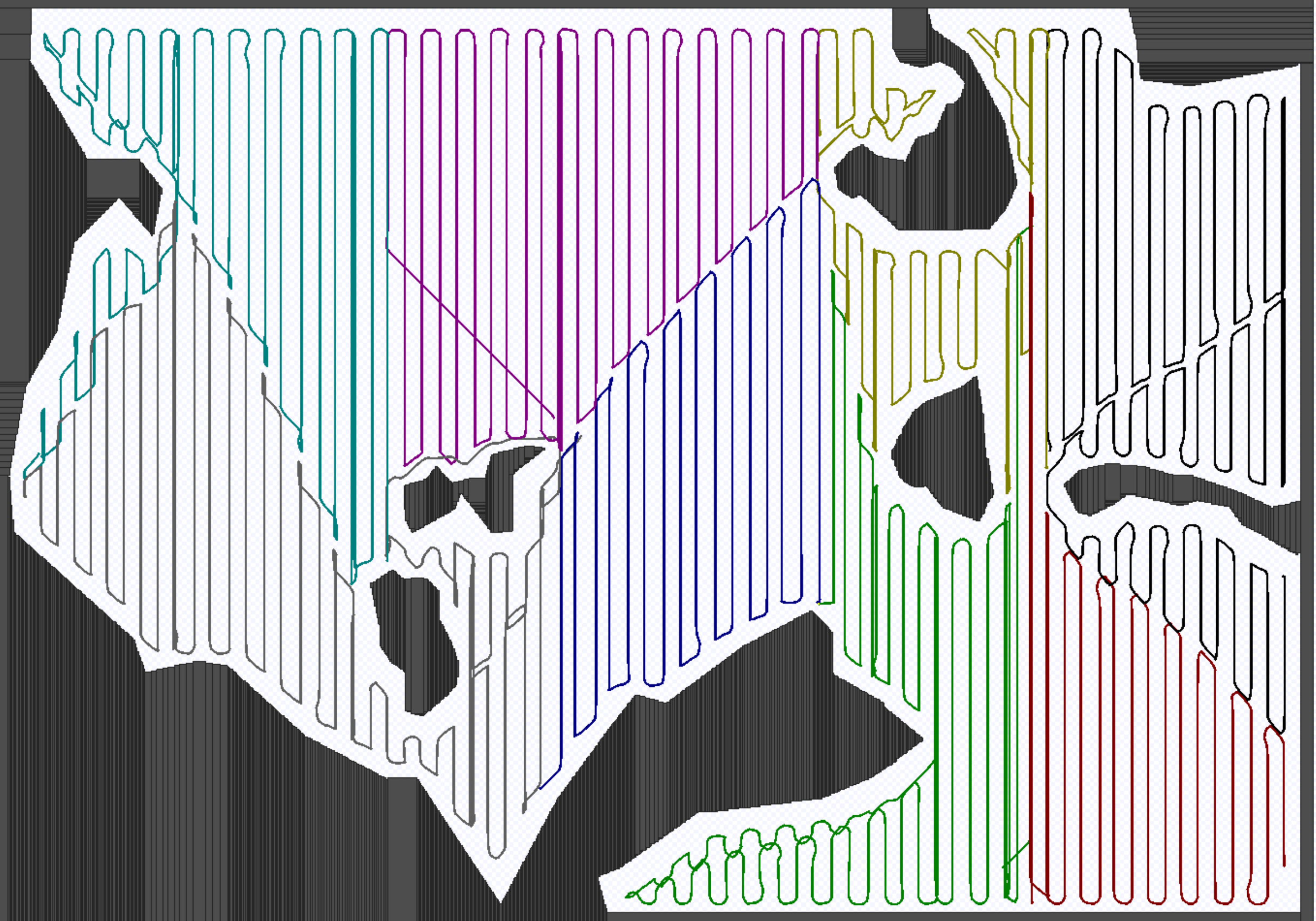}
  \caption{ A team of 8 robots covering a test environment.}
  \label{beauty}
\end{figure}

While for some applications, such as floor sweeping,  scanning of every single point in free space is not crucial, in many situations missing a spot, such as in the case of de\hyp mining operations,  could have catastrophic results.  With this consideration, coverage algorithms can be categorized as either \emph{complete} or \emph{randomized}.  Complete coverage algorithms ensure that each point in the region is visited, while for a randomized approach there is no guarantee on full coverage of the entire area. This work aims to ensure complete coverage with minimum  or no-overlapping regions.

Some coverage algorithms require prior information about the environment. Algorithms that operate on a known area utilizing an existing map are called \textit{off\hyp line} algorithms;  otherwise they are termed \textit{on\hyp line}~\cite{Choset-2001}.  Furthermore, many existing solutions for multi\hyp robot area coverage require constant communication, so the robots can coordinate their tasks.  In many applications, however, it is not possible to guarantee continuous communication. In our work, we address the off-line complete coverage problem by utilizing the cellular decomposition approach.  We further assume that the robots cannot communicate with each other during the operation and that each robot knows the map of the area it is supposed to cover; see Fig.~\ref{beauty}. 

The next section presents an overview of existing solutions for the complete coverage problem for both single and multiple robotic systems. 
Furthermore, we define area coverage in terms of \emph{arc routing} in graph theory and discuss some of its variations. 
In Section~\ref{ch:probFormul}, we present the problem statement and in the next two sections outline the proposed algorithmic solutions. 
Then  experimental results of the graph routing phase and simulated experiments of the multi\hyp robot area coverage algorithms are presented in Section~\ref{sec:results}. Finally, future work is highlighted in Section~\ref{sec:conc} together with a discussion on lessons learned. 

%%%%%%%%%%%%%%%%%%%%%%%%%%%%%%%%%%%%%%%%%%%%%%%%%%%%
\vspace{-0.1in}\section{Related Work}
\label{ch:litReview}

Most approaches to the area coverage problem first decompose the free area in an approach\hyp specific fashion then proceed to compute a global solution. As such, coverage problems can be classified based on the approach taken in the decomposition step. The two most popular approaches are grid\hyp based decompositions and cellular decompositions. 

In grid\hyp based approaches, the area to be covered is represented as a grid map.  Each grid cell's value indicates whether or not it contains an obstacle.  One of the earliest grid\hyp based solutions, introduced by Gabriely and Rimon~\cite{Gabriely-2001}, was building a spanning tree from the map representation. Spanning tree based covergae was adapted for multi\hyp robot systems as well ~\cite{Hazon-MSTC,agmon2008giving, Faziletal.2010, 6608838}. The main challenge of grid\hyp based approaches is that the robustness and efficiency of these methods depends on the resolution of the input representation. As the footprint increases the discretization of the grid becomes too coarse. 

\begin{figure}[t]
      \centering      
      \includegraphics[width=0.35\textwidth]{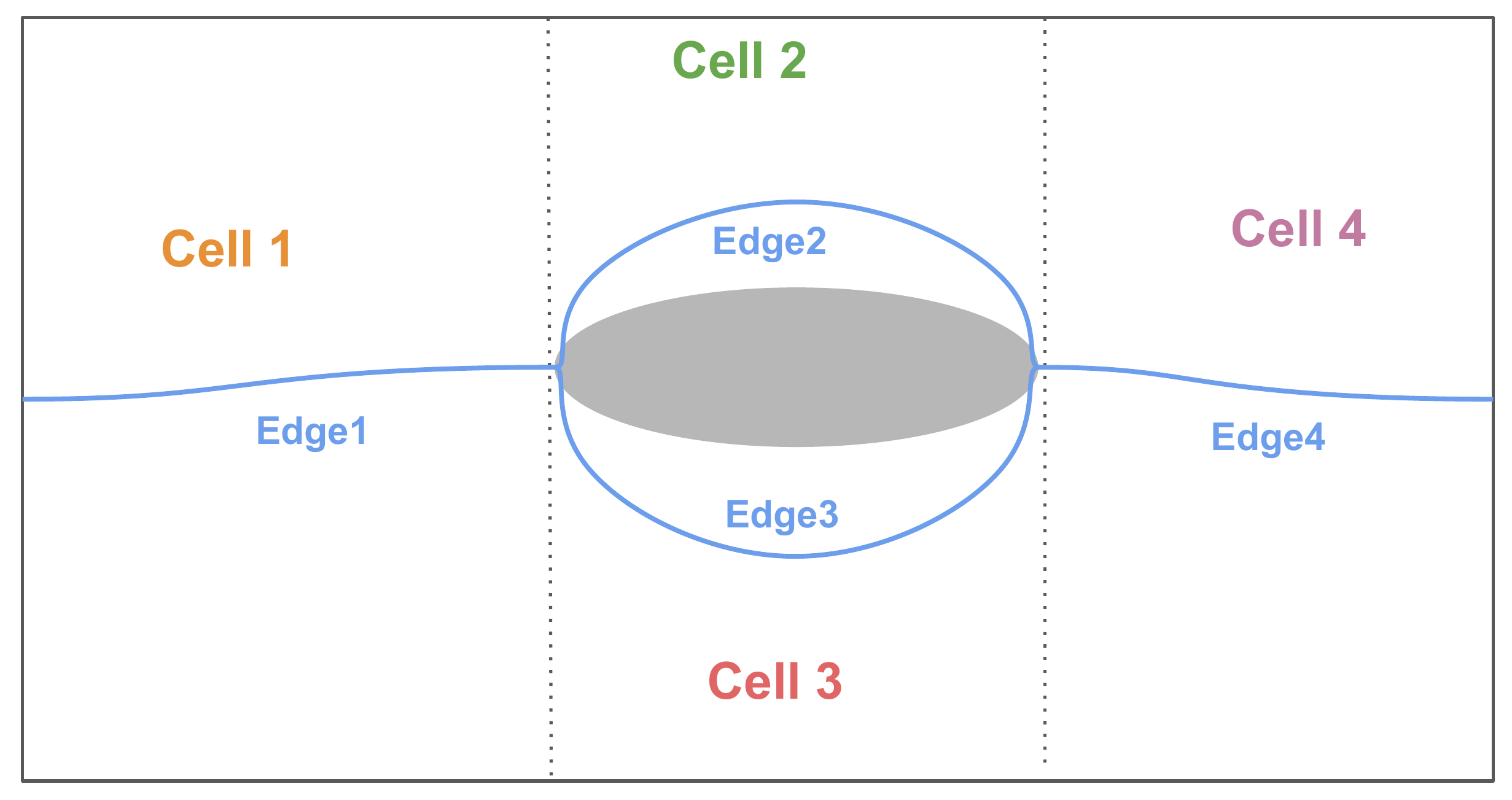}
     \caption{Cellular Decomposition and Corresponding Reeb Graph}
      \label{reebGraph}
\end{figure}

Another popular approach to coverage problems is based on the \emph{cellular decomposition} of the free area. The idea behind cellular decomposition is to split the total free space into non\hyp overlapping cells such that their union is the original area. The resulting decomposition is often represented by an adjacency graph, where either each node is a cell, and the nodes of neighbouring cells are connected by an edge; or each cell is an edge and the points where connectivity changes represent the vertices, the graph then is termed \emph{Reeb graph}; see Fig.~\ref{reebGraph}.
%\invis{
%}

%called 
One of the widely used exact cellular decomposition algorithms is the \emph{boustrophedon decomposition} introduced by Choset in 2000~\cite{Choset-2000,Acar2002J1,Acar2002J2}. The algorithm does not require any prior knowledge of the environment and is designed for single\hyp robot systems.  Each cell in a boustrophedon decomposition is defined by points along obstacles, called \emph{critical points}.  In contrast to other cellular decomposition algorithms, the boustrophedon approach provides a more effective space division.

In 2010, Manadiar and Rekleitis~\cite{Rekleitis2010a} introduced a new efficient algorithm for single robot coverage based on the boustrophedon decomposition. However, in contrast to the original algorithm, the environment is assumed known and the problem is cast as an instance of the \emph{Chinese postman problem (CPP)}:  given a weighted graph, find a minimum\hyp cost cycle that traverses each edge at least once. With this definition, the coverage problem was solved using a polynomial\hyp time algorithm. That approach was extended to the field of aerial robotic coverage producing efficient coverage solutions taking into account real world considerations such as wind direction~{\cite{Xu_et_al.-2014}. This paper extends the efficient coverage algorithm proposed above to multiple robots. 

The work of Rekleitis et al. ~\cite{Rekleitis-2008} extended the boustrophedon decomposition approach from single robots~\cite{Choset-2000} to multi\hyp robot systems for covering  unknown arbitrary environments. Two types of algorithms were presented based on the communication restrictions, called \emph{team\hyp based coverage} and \emph{distributed coverage},  both assuming an unknown environment and utilizing a sensor\hyp based approach.

A recent study by Avellar et al.~\cite{s151127783} on multi\hyp robot coverage that is similar to that addressed by Xu et al.~\cite{Xu_et_al.-2014}  operates in two phases. First the area is decomposed into line\hyp sweeping rows, thus constructing a complete graph where the vertices consist of  the borders of coverage rows  and the launching point. The resulting graph is used in the second phase to solve  the vehicle routing problem~\cite{Toth:2001:VRP}. The authors claim this solution ensures minimum\hyp time coverage for Unmanned Aerial Vehicles (UAV), in contrast to other existing solutions. 
One of the main objectives of this approach is to minimize the number of robots that are required for completing the assigned work.  However, this algorithm is only applicable if there are no obstacles in the area of coverage.

Multi\hyp robot coverage using the environment as an implicit communication device has been proposed under the term {\emph Ant Robotics}; see for example the work by Wagner et al. ~\cite{wagner1999distributed}. More recently repartitioning the area was proposed for covering the space by a team of robots~\cite{hungerford2016repartitioning}. It is worth noting that, the problem of coverage is similar to exploration, especially when considering a much larger\hyp footprint sensor; see for example the work by Quattrini Li et al.~\cite{QUATTRINIPhd,QuattriniLi2016}. Most of the existing studies rely heavily  on the existence of communication between robots, however, in some domains it is infeasible to maintain communications.  To overcome this problem in the agricultural domain, Janani et al.~\cite{Janani:2016:Abst,Janani2016}  proposed a solution that used indirect communication, observing the visual trails of the other robots to guide the coverage.  This kind of a solution is impractical in aerial or marine explorations, as any trail disappears very fast in such environments.

Most of the studies presented the decomposed areas as graphs to perform planning, e.g. graph vertices represent the destination points in the area, and the edges are paths connecting them.  With this graph representation, the optimal area coverage problem can be modelled as a modified version of the \emph{travelling salesman problem (TSP)}  or the  \emph{$k$\hyp travelling salesman problem} for multiple robots. Meanwhile, when the task is to traverse all the edges of a given graph at least once, then the problem is known as the Chinese Postman Problem (CPP) and k\hyp CPP.
While the CPP was proven to have a polynomial\hyp time solution by Edmonds and Johnson~\cite{cpp},  others variants are NP\hyp hard~\cite{Frederickson:1976}.

%%%%%%%%%%%%%%%%%%%%%%%%%%%%%%%%%%%%%%%%%%%%%%%%%%%%
\vspace{-0.1in}\section{Statement of the Problem}
%\chapter{Problem Statement}
\label{ch:probFormul}

In the following section the problem of multi\hyp robot area coverage is formally presented together with relevant definitions. Then two approximation algorithms solving the multi\hyp robot area coverage problem are described.  Please note that the proposed multi\hyp robot algorithms are  based on the single robot efficient area coverage algorithm proposed by Xu et al. \cite{Xu_et_al.-2014}, which we refer to as \textit{Efficient Complete Coverage} algorithm or \textit{ECC} for brevity. 

The multi\hyp robot area coverage problem, with $k$ robots, is the problem of dividing an input area into $k$ non\hyp overlapping sub\hyp areas such that each robot performs the same amount of work while covering its assigned region.

While information exchange between robots may help to avoid overlap \cite{Xu_2011_6905,Rekleitis-2008}, in some real\hyp life operations it is impossible to ensure the availability of communication, either because of sensor range or unavailability of a communication network. This work  discusses the communication\hyp less, complete, and optimal coverage problem for a multi\hyp robot system. The optimality of the coverage is defined as a MinMax problem: minimizing the maximum  coverage cost over a group of robots. The environment that we examine is a 2D bounded rectangular region with obstacles with arbitrary shapes. We assume that there exists prior knowledge about the environment as a map representation. We are also given a single starting point from where all robots must start the coverage and end at that point. In Section \ref{sec:term} we will define some basic terminology used throughout this paper.
\invis{The next two sections \ref{sec:r1c2} \ref{sec:c1r2} will describe the proposed algorithms.}

\subsection{Terminology}
\label{sec:term}

Under the assumption that prior knowledge about the environment exists, the input to the proposed algorithms is a binary image of the coverage area, where obstacles are represented in black and free space in white. The algorithms are also provided with a starting point $v_s$ and the size of the robot team $k$. For the given inputs both algorithms are producing $k$ non overlapping subregions and the tours that connect from the starting point to each subregion and back.

In both algorithms the ECC algorithm is used: from the image map the area is represented as a graph by applying the boustrophedon cellular decomposition (BCD) algorithm~\cite{Choset-2000} and then the CPP algorithm \cite{cpp} is applied to find an optimal route over the complete graph or parts of it. The output of the boustrophedon decomposition is an undirected weighted  Reeb graph $G_r = (V_r, E_r)$: the set of vertices $V_r$ are points on the corners of obstacles that change the connectivity of the region, and the set of edges $E_r$ represent single non\hyp dividable obstacle free cells $C_i$ \cite{Acar2002J2}. We are assuming that a single cell is the smallest unit and thus if the number of cells is less than the number of robots $k$ we will have $E_r - k$ idle robots. The CPP algorithm operates in two stages: it first generates an \emph{Eulerian graph} $G_r$, which is a graph where every vertex has even degree~\cite{Biggs:1986}; then it calculates an \emph{Eulerian path}, which is a path that visits every edge of the graph only one time. An \emph{Eulerian tour} is an Eulerian path that starts and ends at the same vertex.

To create an Eulerian graph some edges in the $G_r$ graph must be duplicated. To perform efficient duplication Xu et al. \cite{Xu_et_al.-2014} define the edge cost  for every edge $e = (v_i, v_j)$ in the $E_r$ as $c_e = \frac{(\text{cell width})^2}{\text{cell area}}$. Additionally,  in this paper, two new weights are introduced to differentiate between the actual coverage cost performed by a robot, and the travel cost  required for traveling  between cells in a subregion. In what follows, the traversal weight $w_t$ of an edge $e=(v_i, v_j)$ is defined as the geodesic distance between the vertices $v_i$ and $v_j$, for all $1 \leq i,j \leq n$; the coverage weight $w_c$ is defined as the size of the corresponding cell area. The starting point is denoted as $v_s = v_1$. The shortest path between vertices $v_i$ and $v_j$ is defined with respect to both traversal and coverage weights: traversal shortest path $sp_t(v_i, v_j)$  and coverage shortest path $sp_c(v_i, v_j)$. The coverage cost of a subregion is defined as the sum of the coverage costs over all cells contained in the subregion and the traversal cost of the paths that connect the cells.

We next present the two $k$\hyp coverage algorithms. 
The first algorithm splits an optimal tour between multiple robots and is called \textit{Coverage with Route Clustering (CRC)}. The second algorithm first divides the area between robots and then starts route planing, and is called \textit{Coverage with Area Clustering (CAC)}.

\section{Coverage with Route Clustering (CRC)}
\label{sec:r1c2}
\begin{figure*}[th]
      \centering
      \fbox{
      \includegraphics[width=0.7\textwidth]{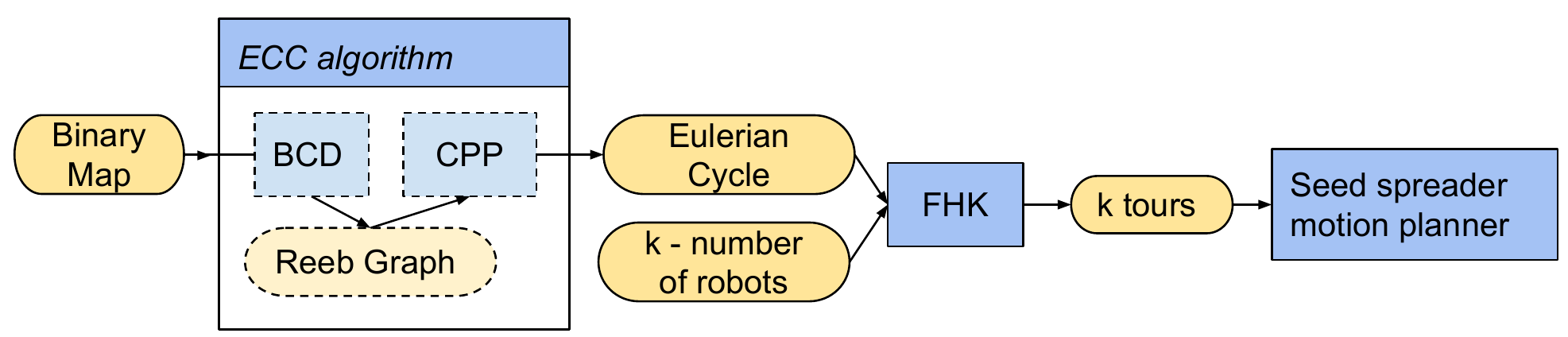}}
     \caption{Coverage with Route Clustering (CRC) execution pipeline}
      \label{pipe1}
\end{figure*}
The first algorithm is a direct extension of the efficient single robot area coverage algorithm (ECC) proposed by Xu et al.~\cite{Xu_et_al.-2014} for a multi\hyp robot system. The pipeline of the algorithm execution is presented in Fig.~\ref{pipe1}. As discussed in the previous section, similar to the original work, the boustrophedon cellular decomposition algorithm  (BCD) ~\cite{Choset-2000} computes a graph representing the free area and encoding the position of obstacles. Because the graph is always connected and all edges are required, the efficient coverage route-finding problem is defined as the MinMax $k$\hyp Chinese postman problem ($k$\hyp CPP). 

For the $k$\hyp CPP problem solution, we adapt the $k$\hyp postman approximation algorithm proposed by Frederikson et al. (FHK) \cite{Frederickson:1976}. The following is a sketch of the algorithm: 
\begin{enumerate}
\item \invis{\\\indent 1.} The efficient Eulerian tour is obtained by the CPP algorithm~\cite{cpp}. 
\item \invis{\\\indent 2.} The most time consuming edge in the Eulerian tour to cover is identified as follows: 
\begin{equation*}
s_{max}=\max\limits_{2 \leq j \leq m-1}\{s(v_1, v_j) + w_c(v_j, v_{j+1}) + s(v_{j+1}, v_1)\}
\end{equation*}
\noindent where $s(v_i, v_j)$ is defined as the shortest path between the vertices $v_i$ and $v_j$ $\forall i, j, 1 \leq i, j \leq n$.
\item \invis{\\\indent 3.} $s_{max}$ is defined as the least amount of work that each robot is assigned to perform. The remaining coverage area is distributed between robots by a fraction of $\frac{j}{k}$.
\end{enumerate}

The FHK algorithm was used to solve the $k$\hyp RPP problem by Xu \cite{Xu_2011_6905}. In their work, there is no differentiation between the coverage and traversal weights of the graph, as the problem domain discussed is different. In contrast to our problem statement, Xu considers road maps, where the coverage and traversal costs can be considered the same. With our problem definition, a modification to the original algorithm is made by introducing the  shortest path $sp_t(v_i, v_j)$ to travel between vertex $v_i$ and vertex $v_j$. The experimental results show that the proposed modification -- using the $sp_t$ cost instead of the $sp_c$ coverage path cost in the computation -- improves the area coverage distribution between robots and decreases the maximum coverage cost. \invis{\textcolor{red}{The experimental results induced in section \ldots}}

\paragraph*{Complexity Analysis} The Exact Cellular decomposition of an area by the BCD algorithm can be performed in $O(n\log n)$ time, where $n$ is the total number of resulting vertices in the free area~\cite{Sleumer99exactcell}. The CPP algorithm finds an Eulerian cycle in polynomial time \cite{cpp}, while FHK algorithm takes $O(n^3)$ time to divide the Eulerian cycle into $k$\hyp tours. Thus, the overall complexity of finding $k$\hyp tours is $O(n^3)$. Although the FHK algorithm might not ensure that all robots are busy,  it provides an optimal solution with a $2 - \frac{1}{k}$ approximation factor \cite{Frederickson:1976} for the MinMax constraints.
\section{Coverage with Area Clustering (CAC)}
\label{sec:c1r2}

The CAC approach is the following: divide the area into approximately equal partitions and apply the Efficient Complete Coverage (ECC) algorithm \cite{Xu_et_al.-2014} for each of the sub\hyp areas. As opposed to the \emph{Cluster First, Route second} algorithm proposed by Xu \cite{Xu_2011_6905}, our algorithm uses a different clustering technique and takes as  input an area map instead of a graph. A complete pipeline of the algorithm is presented in Fig.~\ref{pipe2}.

\begin{figure*}[th]
      \centering
      \fbox{\includegraphics[width=0.7\textwidth]{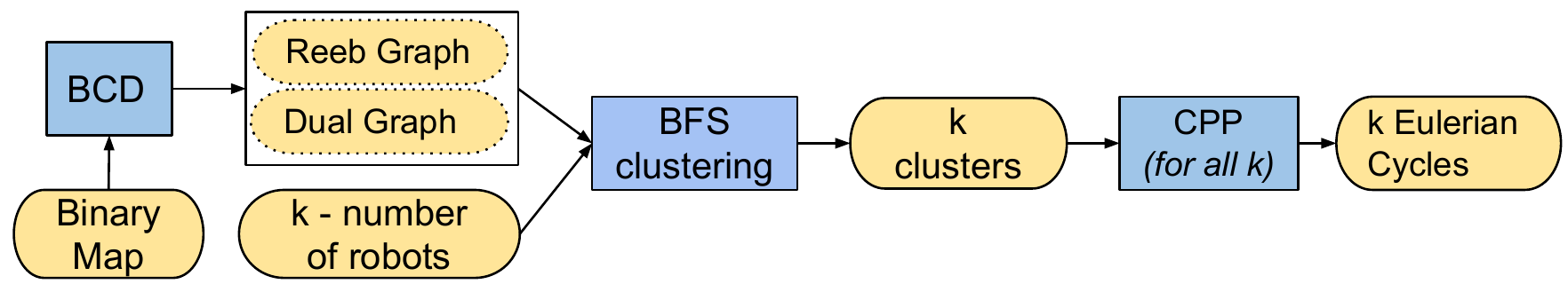}}
     \caption{Coverage with Area Clustering (CAC) execution pipeline}
      \label{pipe2}
\end{figure*}

The proposed algorithm is discussed next, together with the introduction of some additional terminology relevant only to this algorithm.

The \emph{dual graph} of a planar Reeb graph 
%is a graph where each vertex corresponds to an edge of the Reeb graph. n other words, it 
is a weighted graph $G_d = (U_d, E_d)$  with vertex set $U_d = \{u_1, u_2, \ldots, u_m\}$, where each vertex is located at the center of a cell of the Reeb graph;  two vertices in $G_d$ are connected with an edge if and only if their corresponding cells in the Reeb graph are neighbors (share a critical point).  The weight $w_d(u_i, u_j)$ of each edge $(u_i, u_j)$ in $E_d$  is defined as the Euclidean distance between $u_i$ and $u_j$; see Fig.~\ref{DualGraph}. %\invis{; see Figure~\ref{DualGraph}.  Every vertex in ${G_d}$ corresponds to a cell in the Reeb graph $G_r$}. 
As every cell is bounded by two critical points, then we can associate every leftmost critical point with the vertex corresponding to that cell in the dual graph.

%\invis{
\begin{figure}[t]
      \centering      
      \includegraphics[width=0.35\textwidth]{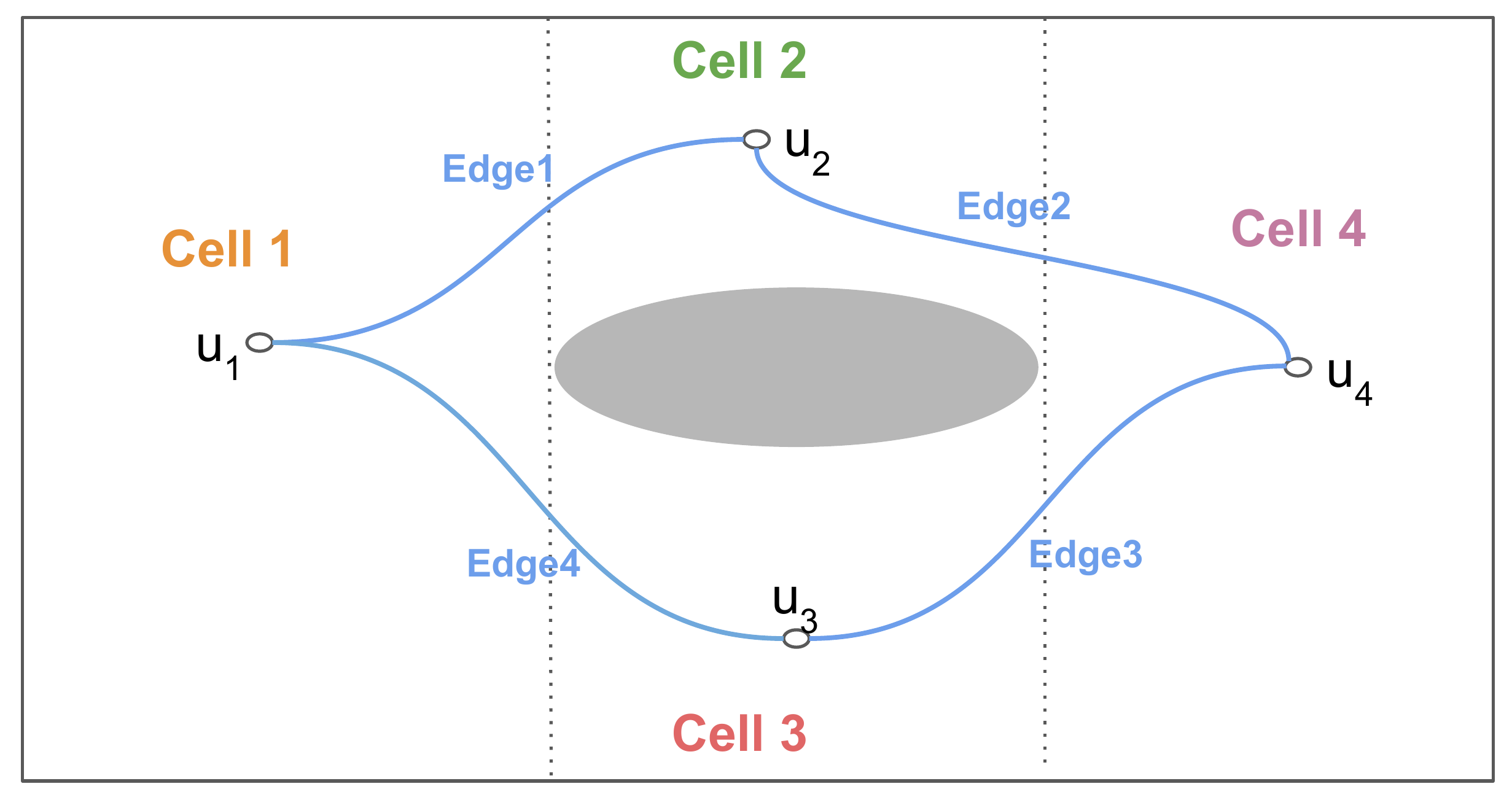}
     \caption{Dual graph representation based on boustrophedon decomposition}
      \label{DualGraph}
\end{figure}
%}

Similar to the CRC approach, the input to the algorithm is an area map that we decompose  into a set of cells  utilizing the BCD algorithm. Along with the Reeb Graph creation, the dual graph $G_d = (U_d, E_d)$  is also built. 
The dual graph is used to divide the area and allocate it to the $k$ robots using a breadth\hyp first\hyp search\hyp like (BFS) algorithm for clustering. We perform a BFS traversal and include every vertex into a cluster as long as the weight limit of the current cluster is not exceeded. 
Two vertices $u_i$ and $u_j$ in the dual graph are considered \textit{nearest neighbors} if and only if 
the Euclidean distance $w_d(u_i, u_j)$ between these vertices has the smallest value for all vertices $u_i,u_j$, where $1 \leq i,j \leq n$. The \textit{cluster size} of a cluster is the sum of the areas of all the cells in the cluster, plus the shortest travel distance from the starting vertex to the nearest cell in the cluster.

\subsubsection*{Area clustering} Clustering of the areas is performed following a greedy approach. 
Each cell in the cluster corresponds to a vertex in the dual graph $G_d$, thus the clustering becomes a vertex clustering problem in a graph. The clustering schema is  the following: 
\begin{enumerate}
\item Sort the vertices/cells in increasing order, first by the $y$\hyp coordinate and then by the $x$\hyp coordinate in $S_{cell}$ array. The sorting has to be stable, to ensure that each vertex is processed from bottom to top and from left to right. Set the current vertex $u_{curr}$ to be the first vertex in the sorted list.
\item Perform breadth\hyp first search starting from $u_{curr}$. 
The order of visiting adjacent vertices is defined by the nearest neighbourhood, i.e. adjacent vertices are processed from nearest to farthest. While traversing each vertex included in the path, the corresponding cell is included in the current cluster as long as the size of the cluster does  not exceed a predefined limit. 
\item The next $u_{curr}$ is the next vertex from $S_{cell}$ that is not included in any cluster, i.e. not marked as visited. Then the process repeats from Step 2.
\end{enumerate}

\subsubsection*{Cluster limit} The limit for each cluster is defined iteratively. The general idea is to compensate the work of the farther\hyp distance travelling robot by assigning the nearest ones more coverage area instead of simply dividing the area into $k$ equal regions. So the procedure for computing the limit is the following:
\begin{enumerate}
\item Initialize $S$ to be equal to the area size. Sort all the vertices in decreasing order in $ST(u_i)$ according to the shortest distance from the source vertex ${v_1}$ to the associated vertex ${v_j}$ of the Reeb graph.
\item For every initial vertex $u_h$  of the cluster h\hyp th, $L = ST(u_h)$. The h\hyp th cluster size must be no greater than $L + \frac{S}{k-h+1}$.
\item Extract from $S$ the cluster size. If the number of clusters has not reached $k$ (the number of robots) and $S$ is greater than $0$, continue to Step 2.
\end{enumerate}

\paragraph*{Compexity analysis}The running time for the cell decomposition stage with the dual graph creation remains $O(n \log n)$ \cite{Sleumer99exactcell}. Area partitioning has a $O(n^2)$ running time, as the greedy approach for cluster expansion goes over every cell only once.  Finally, the running time for finding an Eulerian cycle for every cluster is equal to  $\sum_{i=1}^{k} O(n_i) = O(n)$, where $n_i$ is the size of $i$\hyp th cluster. Thus the overall complexity of the CAC algorithm is $O(n^2)$.

%%%%%%%%%%%%%%%%%%%%%%%%%%%%%%%%%%%%%%%%%%%%%%%%%%%%
\vspace{-0.1in}\section{Experimental Results}
\label{sec:results}
%\chapter{Experimental Results}
The experimental validation of the proposed algorithms consists of two components: first a large number of randomly generated Reeb graphs were tested for analysing the performance of the area partitioning stage of proposed CRC and CAC algorithms. Second, multi\hyp robot simulations were run in stage mobile robot simulator \cite{stage} for different environments to validate statistical results. In both cases, in addition, to show the advantage of our methods, we compare them with simple equal partitioning and original FHK algorithm with only one edge cost. We will refer to last one as FHK. As for simple equal partitioning, we partition a single optimal route using cluster cost equal to $1/k$ of that optimal route cost, where $k$ is the number of robots. We call this naive route clustering and we will henceforth refer to it as NRC.

\subsection{Statistical Analysis}
\label{ch:results}

For statistical analysis we compare the area partitioning phase performance of proposed algorithms, i.e. k\hyp tour splitting FHK algorithm used in CRC algorithm with two different edge costs and DFS-like graph clustering used in CAC. Moreover we show the performance of these algorithms compared to the baseline, i.e. the original FHK algorithm \cite{Frederickson:1976} and NRC. Before presenting the actual results, the testing framework is described in detail.

\begin{figure}[t]
  \begin{center}
    \leavevmode
    \begin{tabular}{cc}
    \subfigure[]{{\includegraphics[width=0.22\textwidth]{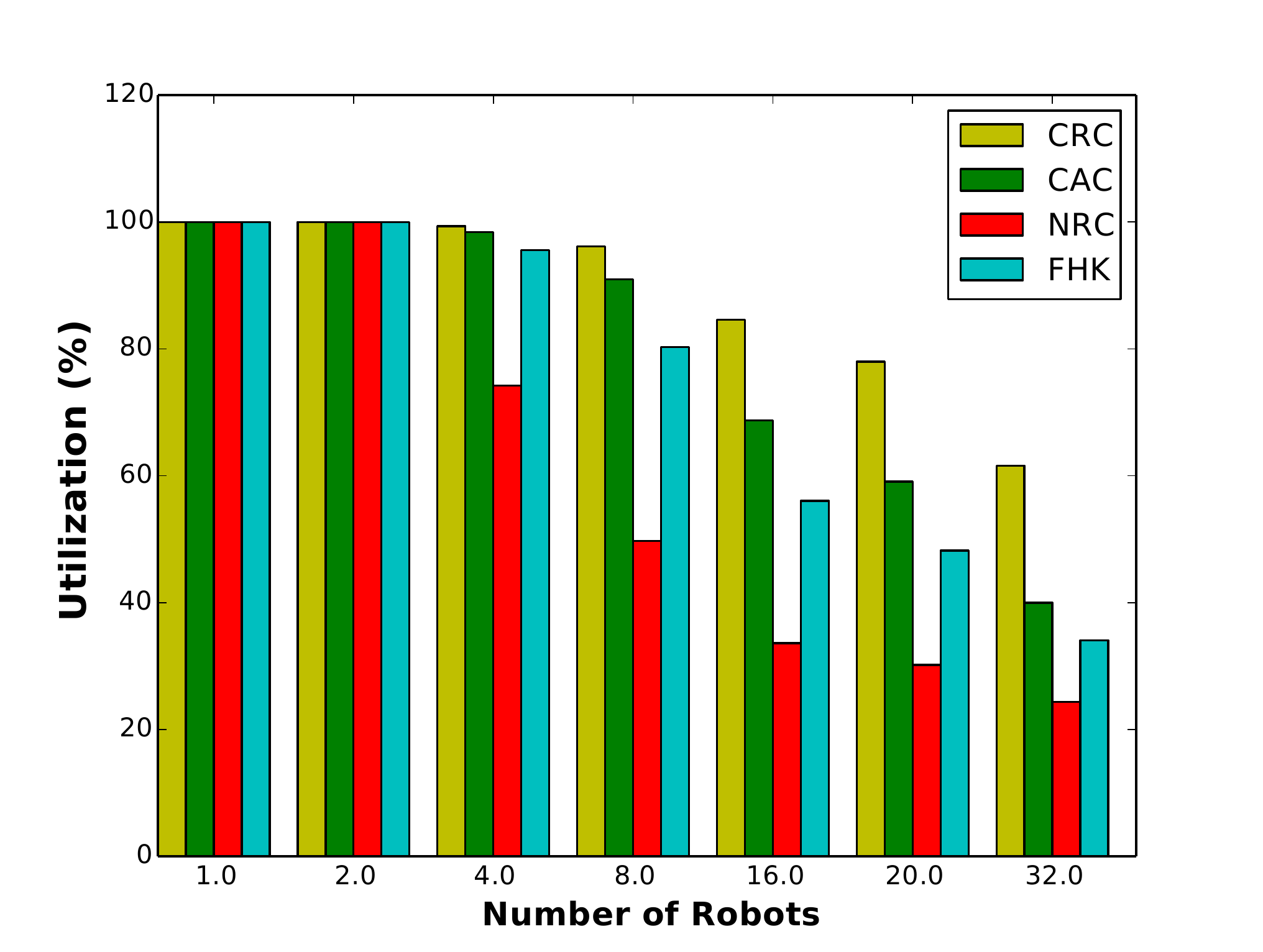}}}&%
    \subfigure[]{{\includegraphics[width=0.22\textwidth]{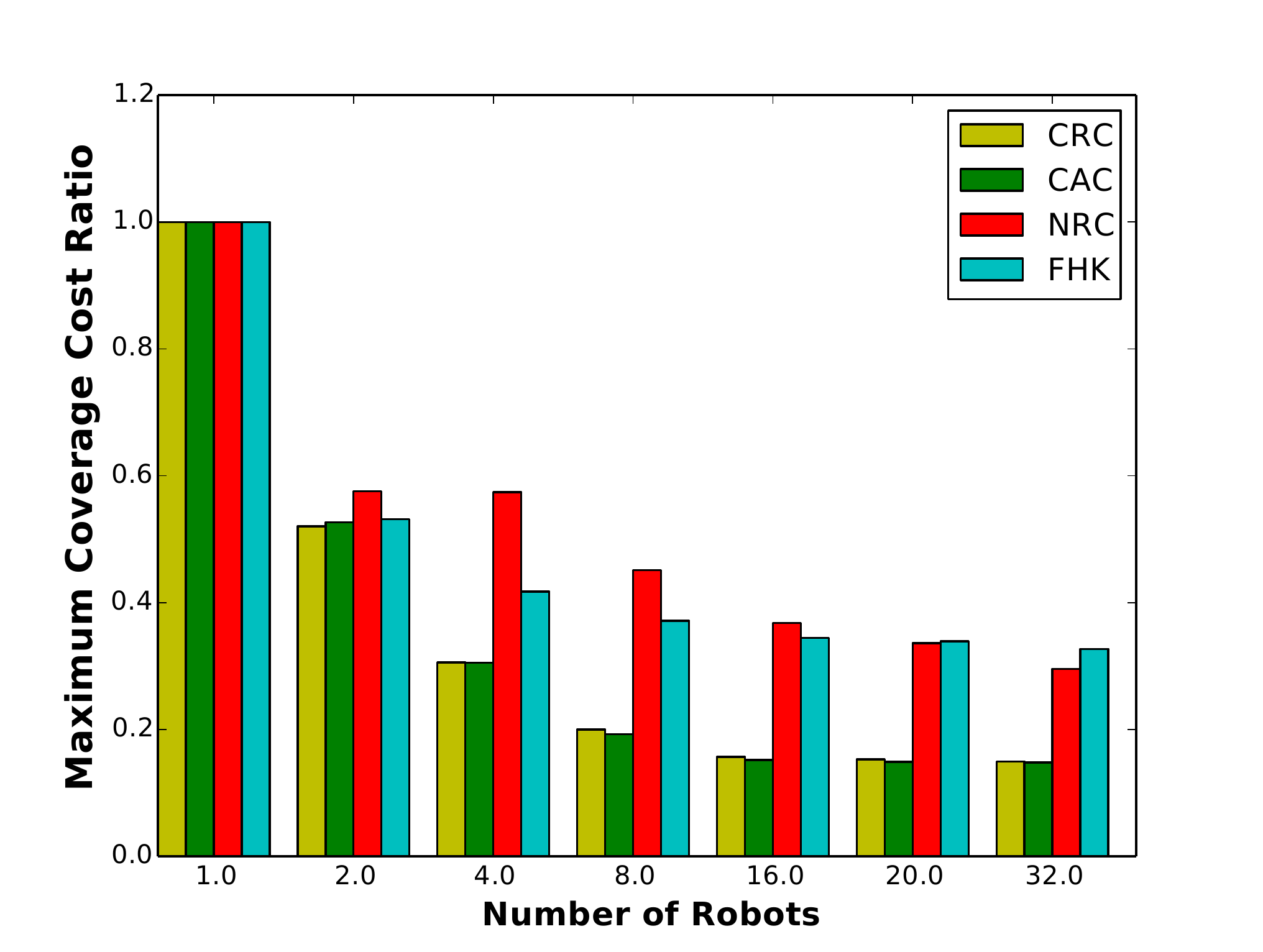}}}%
    \end{tabular}      
  \end{center}\vspace{-0.2in}
 \caption{CRC, CAC  utilization and max coverage cost average results compared to NRC and original FHK algorithms for a variable number of robots. (a) Robot utilization, highest number is better. (b) Maximu area covered, lowest number is better.}
  \label{fig:res}
  \end{figure}

\subsubsection*{Test Data}
Testing data represent randomly generated $200$ images, with arbitrarily distributed obstacles. The boustrophedon decomposition is performed on each of these artificial environments and as a result a Reeb graph is produced. In the resulting graph the Eulerian path is constructed by applying the algorithm solving the Chinese postman problem.  The following input is required for the comparison of the algorithms: a Reeb graph with coverage cost and travel cost assigned to each edge, an Eulerian tour of that graph, and a number of tours the graph is partitioned  into.  Size information about the input graph and the single coverage tour costs are shown in Table \ref{Tab:1}. The number of robots or tours that the graph will be partitioned into have the following values  $ k = \left\{ 1, 4, 8, 16, 20, 32 \right\}$. All data examples are used with all possible $k$ values. It is worth noting that single cell coverage is considered an atomic action. As such, it is only divided where an edge (cell) has to be duplicated (at most once).

\begin{table}[t]
\caption{Input Graphs Information\vspace{-0.2in}}
\label{Tab:1}
\begin{center}
\begin{tabular}{|l|c|c|c|}
\hline
& {\bf MIN}&{\bf MAX}& {\bf MEAN}\\
\hline
Number of Vertices& 10 & 41 &25\\
\hline
Number of Edges & 12 & 57 & 35\\
\hline
Length of Euler Tour & 16 & 73 & 43\\
\hline
Cost of Euler Tour & 2414 & 11176 & 5123\\
\hline
\end{tabular}
\end{center}
\end{table}

\subsubsection*{Measurement Metrics}
For every call of the algorithm, we measure the number of idle robots and the maximum coverage cost. In Fig.~\ref{fig:res} the maximum coverage cost is represented as fraction of single optimal coverage path, and utilization as the percentage of coverer robots. We present average results over all input data for each number of robots Fig.~\ref{fig:res}. In addition to measure overall effectiveness we averaged results over all $k$  robots, when $k>1$ \ref{Tab:2}. 

\subsubsection*{Results}
The area coverage problem is defined in this work as a MinMax problem. The maximum coverage cost is minimized by ensuring that no robot stays idle. However, even with this definition  there is a possibility to have idle robots when the size of cells are not balanced.

\begin{table}[t]
\caption{Average Results for \lowercase{k}=\{2, 4, 8, 16, 20, 32\} robots\vspace{-0.2in}}
\label{Tab:2}
\begin{center}
\begin{tabular}{|l|c|c|}
\hline
& {\bf Utilization ($\%$)}&{\bf Maximum Coverage Cost ratio}\\
\hline
CRC& 86.6 & 0.248 \\
\hline
CAC & 76.2& 0.245 \\
\hline
NRC & 52.0 & 0.433 \\
\hline
FHK & 69.0 & 0.388 \\
\hline
\end{tabular}
\end{center}
\end{table}

All algorithms were applied on the datasets which were generated as described at the beginning of this section. Based on the presented results the CRC and CAC algorithms show similar performance on  solving MinMax problem and on utilizing the robots; see Fig \ref{fig:res}. But on average for larger number of robots CRC compared to CAC shows 10$\%$ better utilization. Meanwhile both CRC and CAC algorithms outperform NCR and FHK. In average CRC and CAC outperform NRC and FHK algorithms by maximizing the utilization by 20.5$\%$ and minimizing the coverage cost by 39.8$\%$; see Table \ref{Tab:2}.

\subsection{Experimental Validation in different environments}
For testing complete pipeline of CRC and CAC algorithms different environments were used. In Fig.~\ref{fig:env} the top row presents a variant of the well known cave environment from stage~\cite{stage}; see Fig.~\ref{fig:env}(a). A complex environment with many obstacles is presented in Fig.~\ref{fig:env}(b). An indoor environment is shown in Fig.~\ref{fig:env}(c), and the large environment from rural Quebec from the work of Xu et al.~\cite{Xu_et_al.-2014} is presented in Fig.~\ref{fig:env}(d). The second row of Fig.~\ref{fig:env} presents the coverage path for four robots performing the CRC coverage algorithm, while the third row of Fig.~\ref{fig:env} presents the respective coverage paths utilizing the CAC algorithm. Each of the robots utilizes the Boustrophedon coverage pattern proposed by Acar et al.~\cite{Acar2002J1,Acar2002J2}. As can be seen from the different results in the different environments, the distribution of areas among the robots vastly varies.  To a large extent, the fact that each cell represents an atomic coverage action it is responsible for the uneven distribution of tasks. 

\begin{figure*}[t]
  \begin{center}
  {
 \scalebox{0.8}{
    \leavevmode
    \begin{tabular}{cccc}
      \subfigure[]{\fbox{\includegraphics[height=.125\textheight]{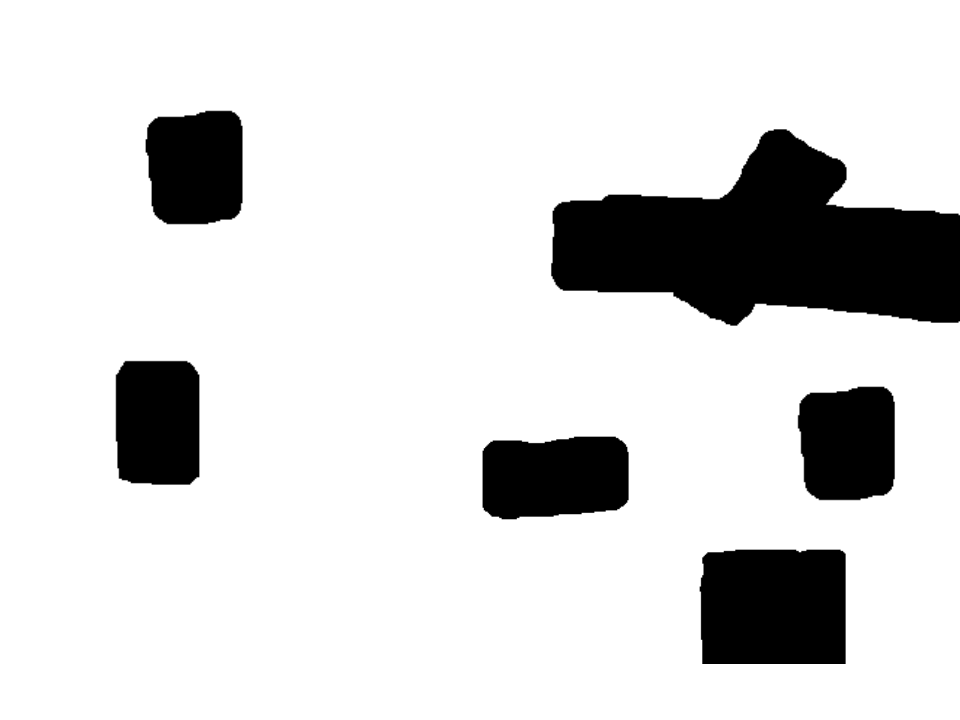}}}\label{fig:ads1}&
      \subfigure[]{\fbox{\includegraphics[height=.125\textheight]{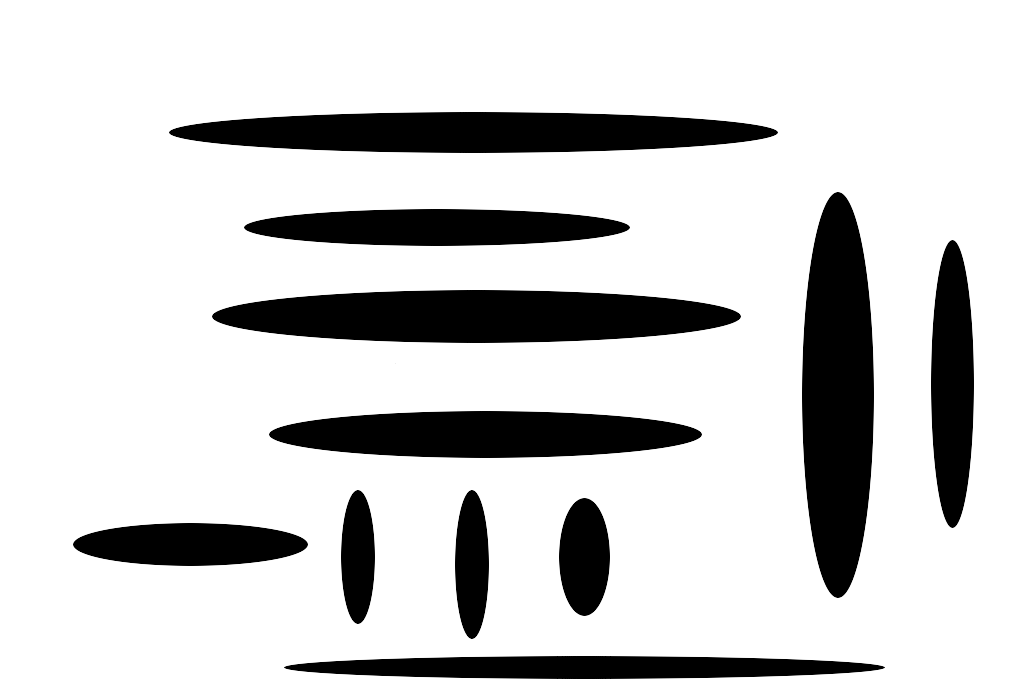}}}\label{fig:ads2}&
     \subfigure[]{ \fbox{\includegraphics[height=.125\textheight]{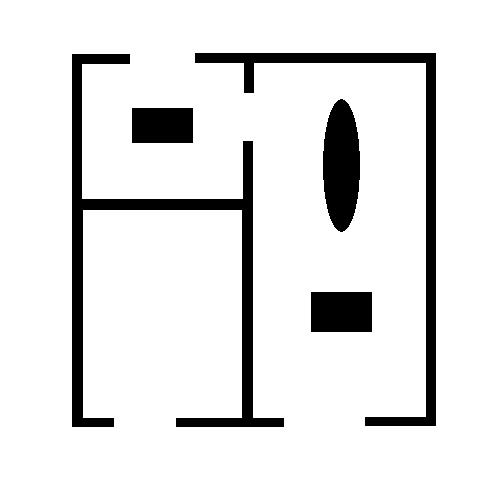}}}\label{fig:ads3}&
      \subfigure[]{\fbox{\includegraphics[height=.125\textheight]{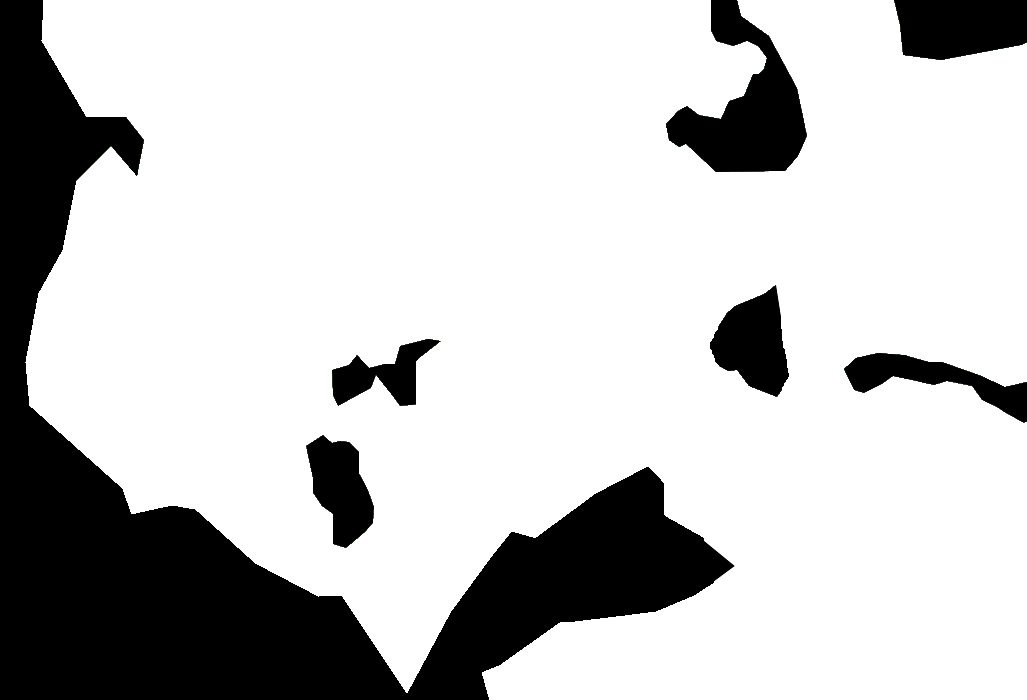}}}\label{fig:ads4}\\
      \subfigure[]{\fbox{\includegraphics[height=.125\textheight]{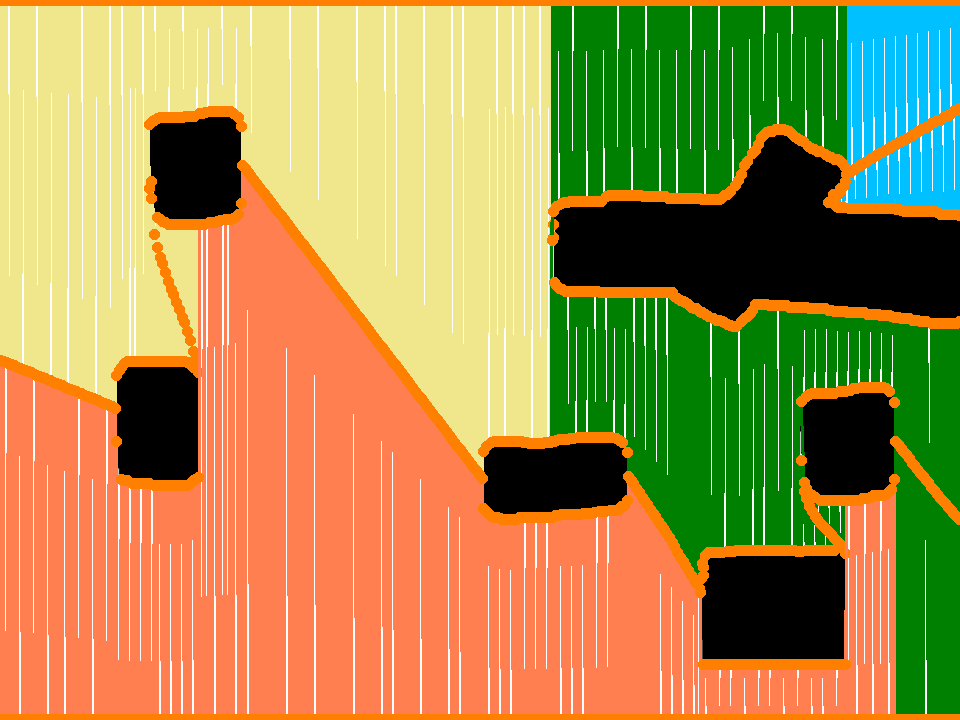}}}\label{fig:ads5}&
     \subfigure[]{ \fbox{\includegraphics[height=.125\textheight]{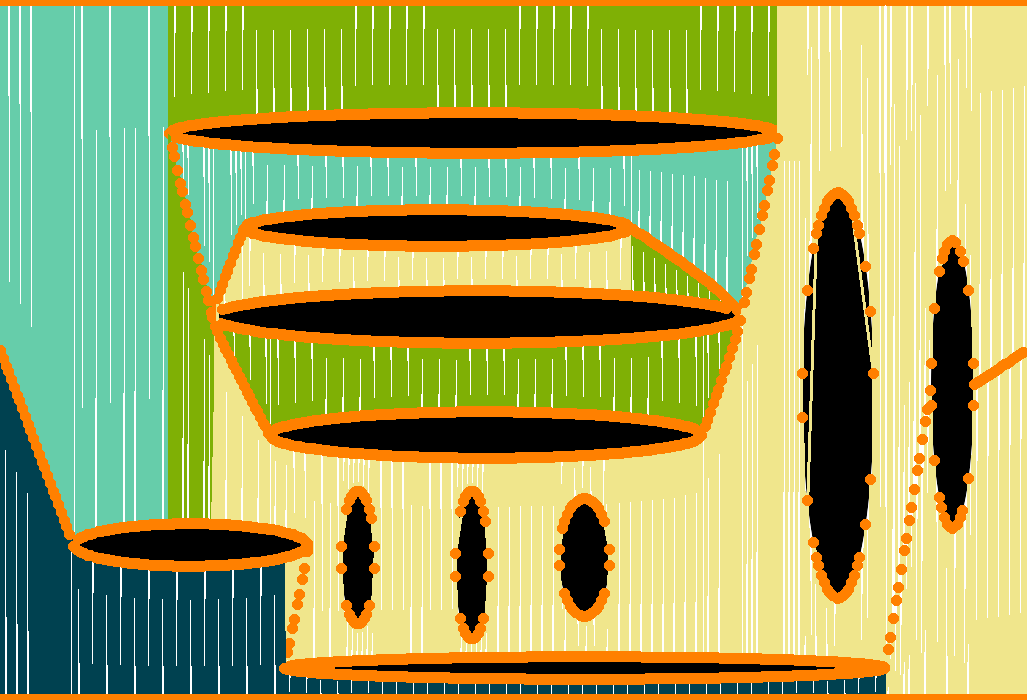}}}\label{fig:ads6}&
      \subfigure[]{\fbox{\includegraphics[height=.125\textheight]{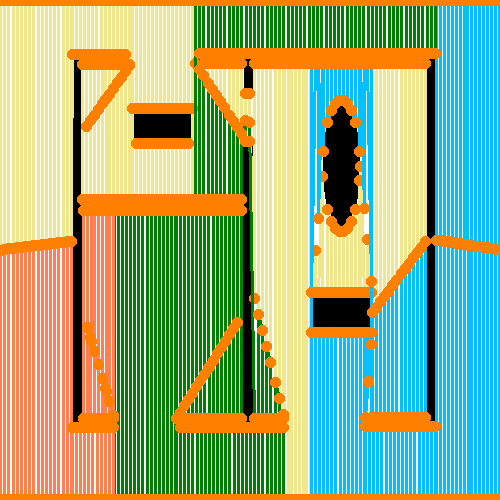}}}\label{fig:ads7}&
      \subfigure[]{\fbox{\includegraphics[height=.125\textheight]{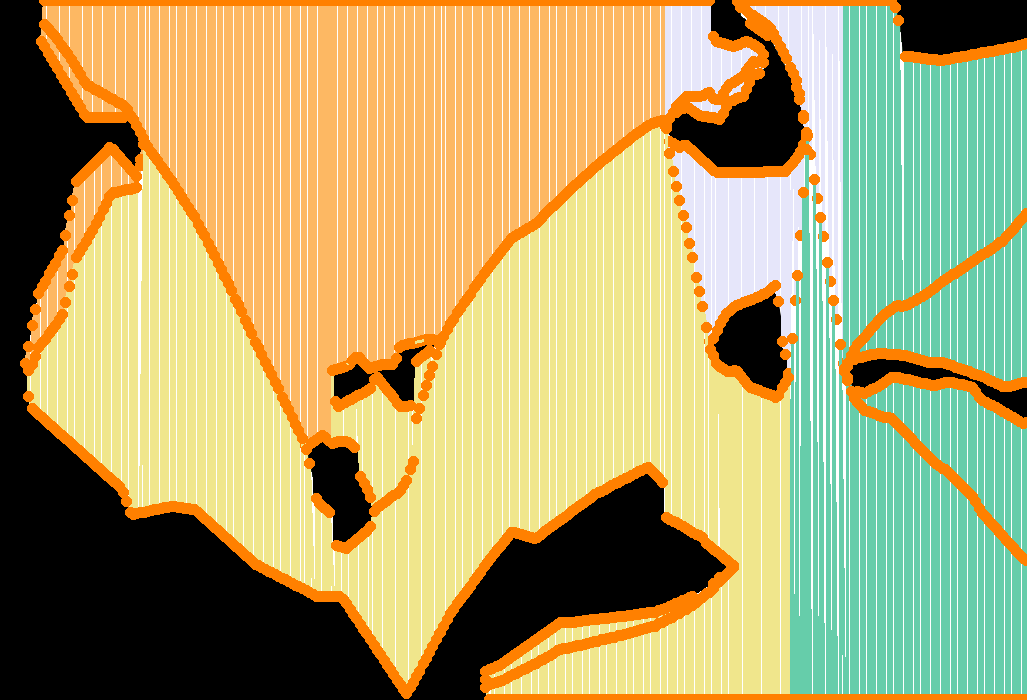}}}\label{fig:ads8}\\
      \subfigure[]{\fbox{\includegraphics[height=.125\textheight]{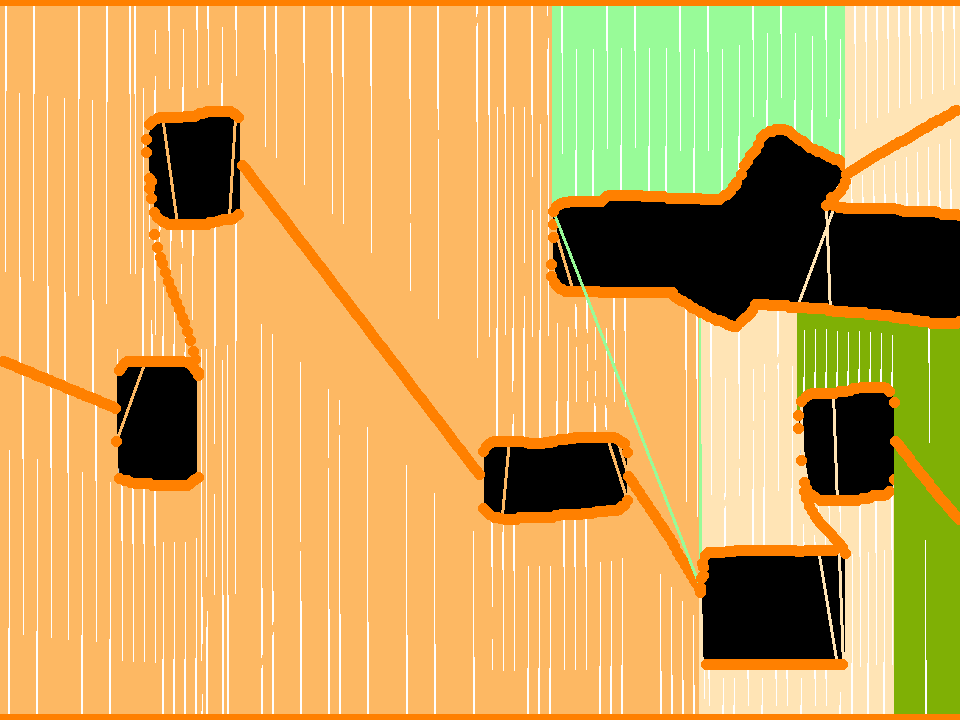}}}\label{fig:ads9}&
     \subfigure[]{ \fbox{\includegraphics[height=.125\textheight]{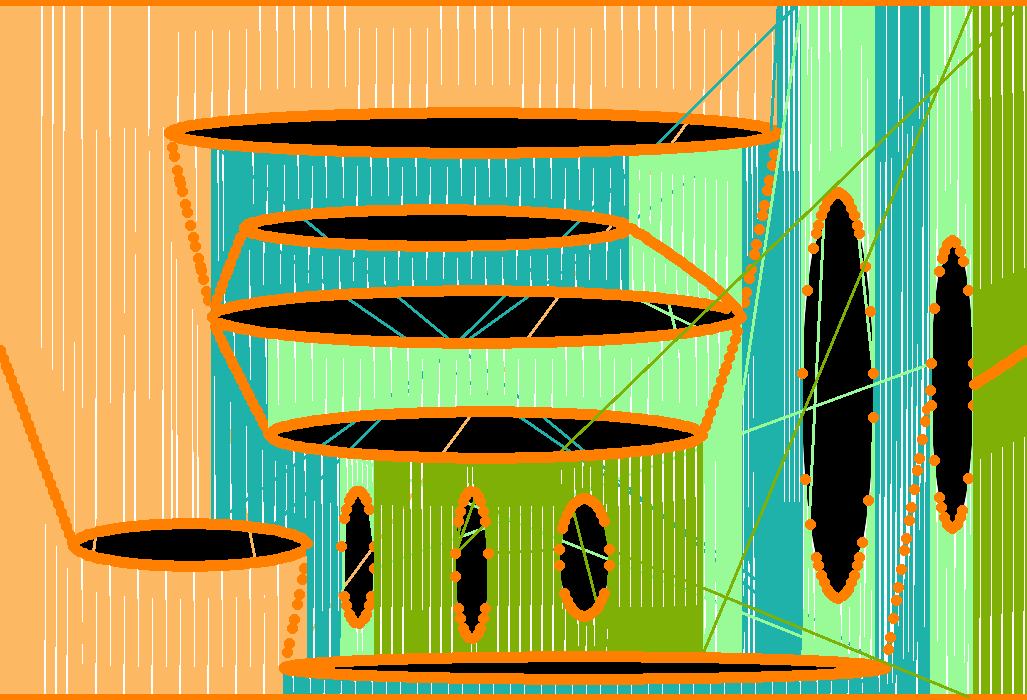}}}\label{fig:ads10}&
      \subfigure[]{\fbox{\includegraphics[height=.125\textheight]{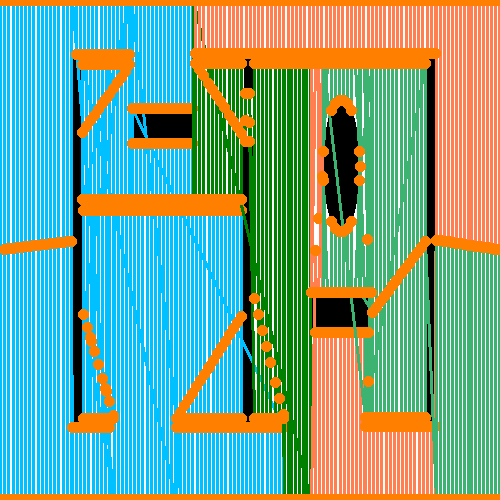}}}\label{fig:ads11}&
      \subfigure[]{\fbox{\includegraphics[height=.125\textheight]{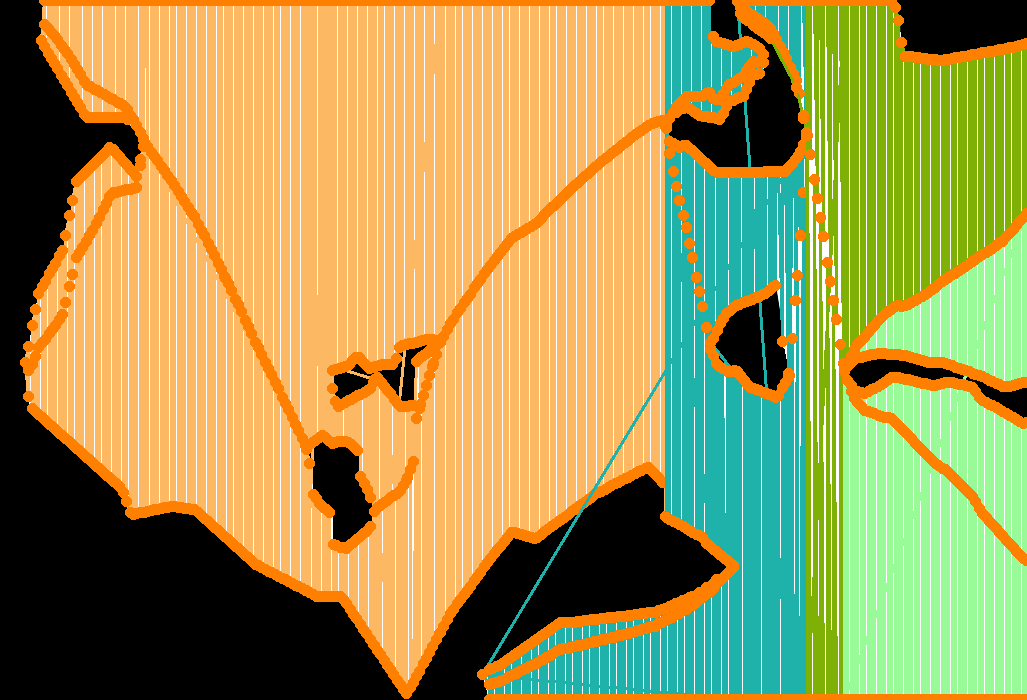}}}\label{fig:ads12}\\
    \end{tabular}      
    }}
  \end{center}\vspace{-0.2in}
  \caption{The four environment where the proposed multi\hyp robot coverage algorithms were tested. (a) Cave; (b) Multi\hyp cell; (c) Indoor; (d) Rural Quebec. (e-h) Coverage paths for four robots utilizing the CRC coverage algorithm. (i-l) Coverage paths for four robots utilizing the CAC coverage algorithm. }
  \label{fig:env}
\end{figure*}

To validate statistical results same four environments were also used to test the performance of the proposed algorithms in comparison with the NRC, FHK algorithms for different number of robots. Fig.~\ref{fig:stats} presents the utilization and maximum coverage cost, calculated the same way as it was described in the previous section,  for covering each of the environments presented in Fig.~\ref{fig:env} (a)-(d). Different number of robots (\{1,2,4,8,16,20,32\}) were used. The results once again show that both CRC and CAC in comparison with naive clustering (NRC) and FHK provide better minimization and utilization, even though they have the same convergence for minimizing the maximum coverage cost. It is worth noting that the scale of each environment is arbitrary and the distances measured only serve as relative measurements. What is significant, however, is the comparison among different number of robots in a single environment. 
 
 \begin{figure*}[t]
  \begin{center}
    \leavevmode
    \begin{tabular}{cc}
    \subfigure[]{\fbox{\includegraphics[width=0.4\textwidth, clip=true, trim=0.5in 0.0in 0.0in 0.0in ]%2.7in 0.0in 2.5in ]
  {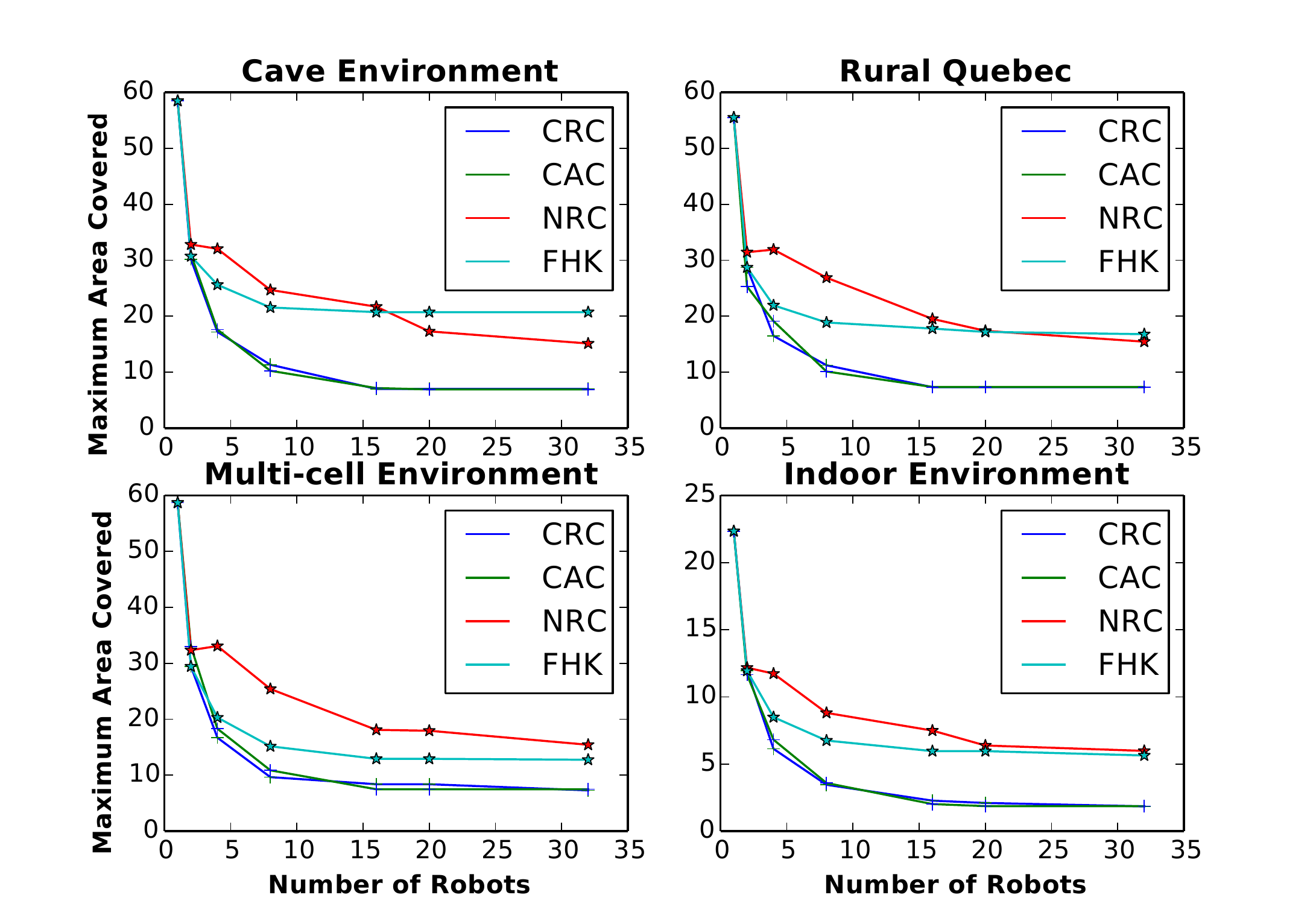}}}&%
    \subfigure[]{\fbox{\includegraphics[width=0.4\textwidth, clip=true, trim=0.5in 0.0in 0.0in 0.0in ]%2.7in 0.0in 2.5in ]
  {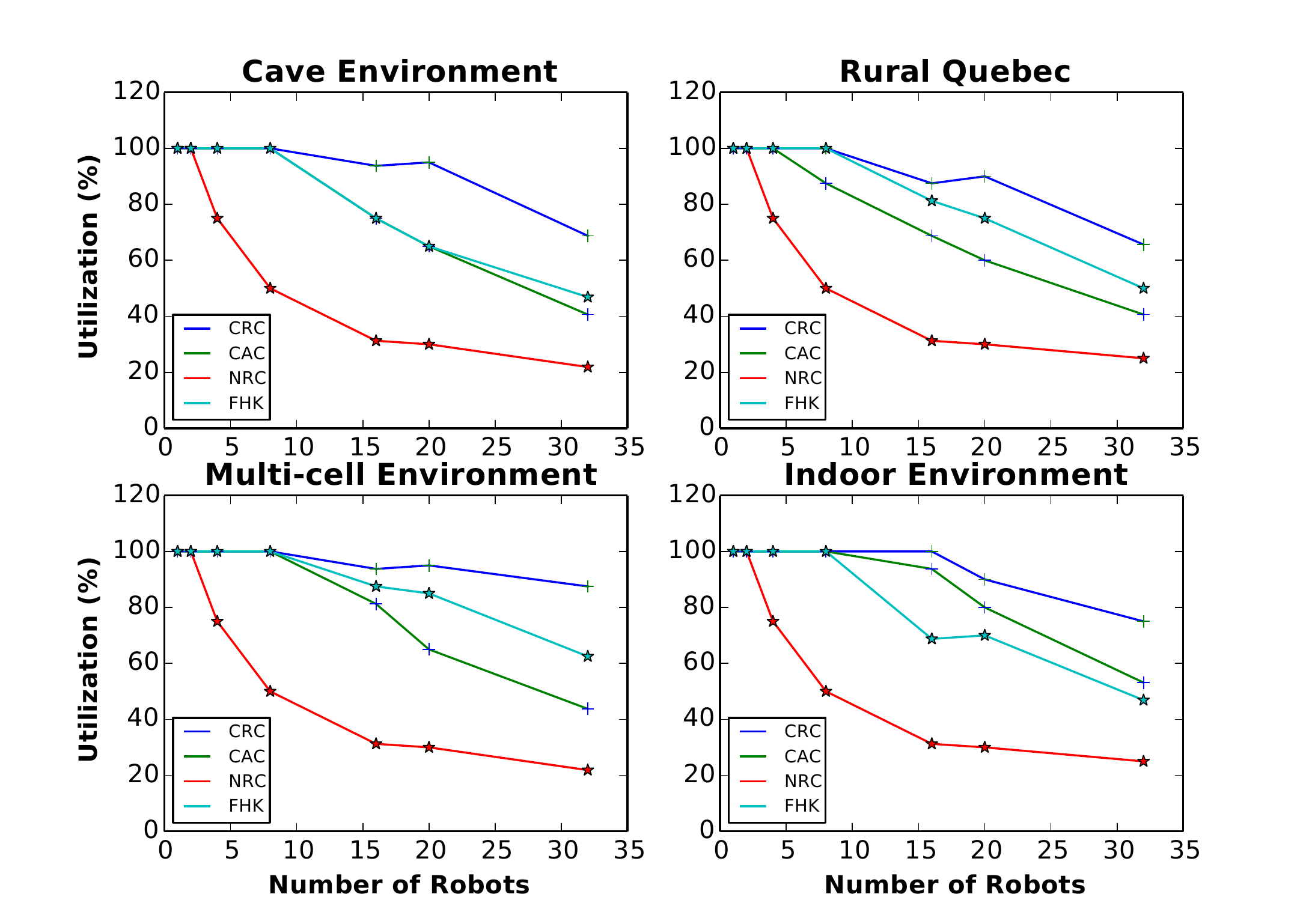}}}%
    \end{tabular}      
  \end{center}\vspace{-0.2in}
 \caption{The effectiveness of the multi\hyp robot coverage algorithms for different environments and for different number of robots.}
  \label{fig:stats}
  \end{figure*}

%%%%%%%%%%%%%%%%%%%%%%%%%%%%%%%%%%%%%%%%%%%%%%%%%%%%
\vspace{-0.1in}\section{Discussion and Conclusion}
\label{sec:conc}
Two solutions to the multi-robot area coverage problem were presented. The optimal coverage in this paper is formulated as a MinMax optimization problem. The ability to communicate between robots was excluded in the formulation of our problem to ensure that the problem is applicable to environments with minimal or no communication.

In the first CRC algorithm, the  ECC algorithm was used to calculate single optimal tour that covers entire area of free space. On the resulting tour, a modification of the  $k$\hyp tour splitting FHK algorithm---termed CRC coverage algorithm---is applied. We show that defining the longest distance in the FHK algorithm,  by taking into account the travel cost of edges instead of just the coverage cost, improves the original algorithm. The running time of the tour splitting stage is $O(V^3)$, with an approximation factor of $2 - \frac{1}{k}$ \cite{Frederickson:1976}. A second greedy approximation algorithm  CAC  with $O(V^2)$ complexity is also presented. It first splits the free space in smaller areas and then performs efficient single robot coverage on each of the cells.

Presented results for the CRC and CAC algorithms illustrate the improvement in robot utilization and minimization of max coverage cost. Even though the proposed algorithms show good performance, they still lack high utilization for a larger number of robots. The latter is due to the unbalanced cell sizes. Thus our future work will study new ways of partitioning the area to achieve better utilization for a larger number of robots. Furthermore, we are currently constructing a fleet of six Autonomous Surface Vehicles (ASVs). The proposed algorithms will be deployed on the ASVs to perform bathymetric studies in the nearby Lake Murray SC, USA. 

%%%%%%%%%%%%%%%%%%%%%%%%%%%%%%%%%%%%%%%%%%%%%%%%%%%%

%\subsection{Figures and Tables}

%Positioning Figures and Tables: Place figures and tables at the top and bottom of columns. Avoid placing them in the middle of columns. Large figures and tables may span across both columns. Figure captions should be below the figures; table heads should appear above the tables. Insert figures and tables after they are cited in the text. Use the abbreviation ÒFig. 1Ó, even at the beginning of a sentence.


%%%%%%%%%%%%%%%%%%%%%%%%%%%%%%%%%%%%%%%%%%%%%%%%%%%%%%%%%%%%%%%%%%%%%%%%%%%%%%%%

%%%%%%%%%%%%%%%%%%%%%%%%%%%%%%%%%%%%%%%%%%%%%%%%%%%%%%%%%%%%%%%%%%%%%%%%%%%%%%%%

%%%%%%%%%%%%%%%%%%%%%%%%%%%%%%%%%%%%%%%%%%%%%%%%%%%%%%%%%%%%%%%%%%%%%%%%%%%%%%%%
%\section*{APPENDIX}


\vspace{-0.1in}\section*{ACKNOWLEDGMENT}

The authors would like to thank the generous support of the Google Faculty Research Award and the National Science Foundation grants (NSF 1513203, 1637876).

%%%%%%%%%%%%%%%%%%%%%%%%%%%%%%%%%%%%%%%%%%%%%%%%%%%%%%%%%%%%%%%%%%%%%%%%%%%%%%%%

%References are important to the reader; therefore, each citation must be complete and correct. If at all possible, references should be commonly available publications.

\bibliographystyle{IEEEtran}
\bibliography{IEEEabrv,refs}

\end{document}